\documentclass[11pt]{article}

\usepackage{microtype} 
\usepackage{booktabs}  
\usepackage{url}  
\usepackage{wrapfig}
\usepackage{graphicx}
\usepackage{multirow}
\usepackage[subtle]{savetrees}
\usepackage[numbers]{natbib}

\usepackage{amsmath}
\usepackage{amsthm}

\usepackage[textsize=tiny]{todonotes}

\newcommand\blfootnote[1]{%
  \begingroup
  \renewcommand\thefootnote{}\footnote{#1}%
  \addtocounter{footnote}{-1}%
  \endgroup
}

\usepackage[final]{automl}
\usepackage{natbib}
\definecolor{Mulberry}{rgb}{0.77,0.29,0.55}

\title{ONNX-Net: Towards Universal Representations and Instant Performance Prediction for Neural Architectures}

\author[1]{\nameemail{Shiwen Qin$^\ast$}{shiwen.qin@ed.ac.uk}}
\author[2]{\nameemail{Alexander Auras$^\ast$}{alexander.auras@uni-siegen.de}}
\author[1]{\nameemail{Shay B.~Cohen}{scohen@inf.ed.ac.uk}}
\author[1]{\nameemail{Elliot J.~Crowley}{elliot.j.crowley@ed.ac.uk}}
\author[2]{\nameemail{Michael Moeller}{michael.moeller@uni-siegen.de}}
\author[3]{\nameemail{Linus Ericsson$^\dagger$}{linus.ericsson@glasgow.ac.uk}}
\author[2]{\nameemail{Jovita Lukasik$^\dagger$}{jovita.lukasik@uni-siegen.de}}

\affil[1]{University of Edinburgh}
\affil[2]{University of Siegen}
\affil[3]{University of Glasgow}

\hypersetup{%
  pdfauthor={Shiwen Qin, Alexander Auras, Shay B. Cohen, Elliot J. Crowley, Michael Moeller, Linus Ericsson, Jovita Lukasik},
  pdftitle={ONNX-Net: Towards Universal Representations and Instant Performance Prediction for Neural Architectures},
  pdfsubject={ONNX-Bench, a unified benchmark across diverse NAS search spaces, and ONNX-Net, a universal text-based representation and surrogate that enable fast performance prediction of arbitrary neural architectures.},
  pdfkeywords={Neural Architecture, Neural Architecture Representation, Task Performance Evaluation}
}

\usepackage{cleveref}

\begin{document}

\maketitle

\begin{abstract}
Neural architecture search (NAS) automates the design process of high-performing architectures, but remains bottlenecked by expensive performance evaluation. Most existing studies that achieve faster evaluation are mostly tied to cell-based search spaces and tailored graph encodings,
limiting their flexibility and scalability when applied to more expressive search spaces. 
In this work, we aim to close the gap of individual search space restrictions and search space-dependent network representations. We present ONNX-Bench, a benchmark consisting of a collection of neural networks in a unified format based on the Open Neural Network Exchange (ONNX) format. ONNX-Bench\footnote{\url{https://huggingface.co/datasets/carlosqsw/ONNX-Bench}} 
includes all open-source NAS-bench-based neural networks, resulting in a total size of $650$k $\{$architecture, accuracy$\}$  pairs. Using this benchmark, we create a shared neural network representation, ONNX-Net, able to represent any neural architecture using natural language descriptions acting as an input to a performance predictor. This text-based encoding can accommodate arbitrary layer types, operation parameters, and heterogeneous topologies, enabling a single surrogate to generalise across all neural architectures rather than being confined to cell-based search spaces. Experiments show strong zero-shot performance across disparate search spaces using only a small amount of pretraining samples, enabling the unprecedented ability to evaluate any neural network architecture instantly. Our code is available at \url{https://github.com/shiwenqin/ONNX-Net}.
\end{abstract}
\blfootnote{$^\ast$ Equal contribution.}
\blfootnote{$^\dagger$ Equal contribution.}


\section{Introduction}
Neural Architecture Search (NAS) aims to automate neural network design, yet it has largely struggled to discover fundamentally new architectures. One contributing factor is the use of restrictive search spaces, like cell-based spaces, which limit exploration to a single class of network designs \citep{ying2019bench, ChenPFL21}. While more expressive search spaces have recently emerged \citep{schrodi2023construction, ericsson2024einspace}, they exacerbate the primary bottleneck of NAS: the high computational cost of architecture evaluation.

Surrogate models \citep{dudziak2020brp, lee2021hardware} mitigate this cost by approximating the evaluation function by learning a mapping from architecture representations to their evaluation performance. Early surrogate models \citep{tang2020semi, wen2020neural} rely on fixed search space priors, making them difficult to transfer across diverse spaces. More recent approaches \citep{mills2023gennape, akhauri2024encodings} aim at improving generalisation across search spaces. 
GENNAPE \citep{mills2023gennape} is able to represent any neural network using computational graphs encodings of operation bundles. FLAN \citep{akhauri2024encodings} uses adjacency-matrix encodings.  
However, these approaches have important limitations. First, adjacency-matrix encodings scale well only in cell-based spaces, and struggle to extend to more flexible search spaces with sparse variable-sized graphs. Second, graph-based structures are largely insensitive to operator parameterisation: two architectures with identical graphs but different operator parameters (e.g., convolution kernel size) are often indistinguishable. This underlines the strong need for a representation that captures both topology and rich operator-level details in a search space-agnostic manner.

\begin{figure*}
    \centering
    \includegraphics[width=0.85\linewidth]{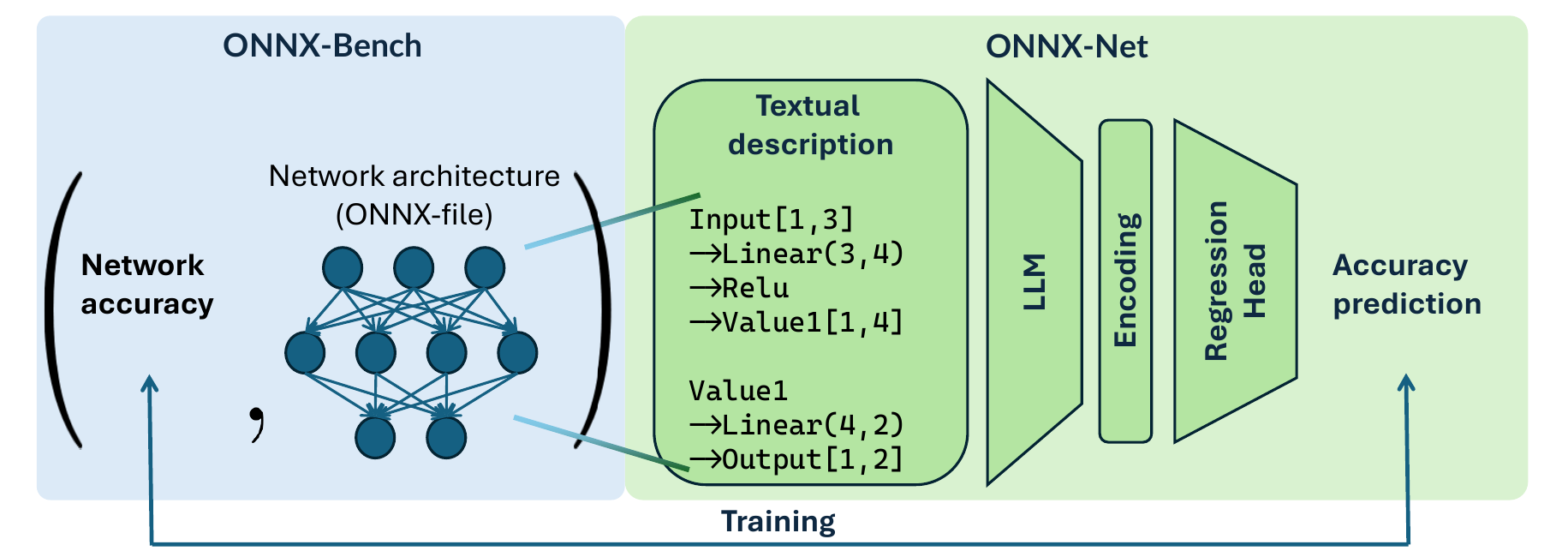}
    \caption{Overview of our approach: ONNX-Bench contains $\{$architecture, accuracy$\}$ pairs from multiple search spaces in unified ONNX representation; ONNX-Net consists of a robust, universal text encoding and an LLM-based performance predictor.}
    \label{fig:contrib_overview}
\end{figure*}

To address this, we develop a more general and robust encoding for NAS surrogate models, \textbf{ONNX-Net} (cf. \cref{fig:contrib_overview} (right)) --- one that is agnostic to search spaces, sensitive to operator-level details, and simple to extend.
We propose representing architectures as text generated from Open Neural Network Exchange (ONNX) based computational network representations \citep{onnx} and show that training an end-to-end LLM-based predictor on the text encoding allows for instant performance prediction. 

Text encodings are flexible and compositional; they can naturally capture topology, operators, and fine-grained parameters, as well as auxiliary context, without redesigning the encoder for each space. However, learning effective predictors from text requires a sufficiently diverse collection of architectures spanning multiple search spaces. Existing NAS benchmarks fall short in this regard, as they are typically confined to a single network type and cannot support training predictors that generalise beyond it. To overcome this limitation, we introduce \textbf{ONNX-Bench} (cf. \cref{fig:contrib_overview} (left)), a benchmark that consolidates architectures from multiple search spaces into a unified ONNX-based representation with a consistent evaluation setup. ONNX-Bench provides the diversity needed for training and testing cross-space predictors, and serves as a foundation for studying encodings that capture both structural and operator-level details, such as our proposed ONNX-Net. In our experiments, we demonstrate that a surrogate model using the novel text-based encoding trained on ONNX-Bench achieves competitive performance, especially for zero-shot transferability with minimal pretraining.

Our contributions are as follows:
\begin{itemize}
    \item We release \textbf{ONNX-Bench}, a large-scale, open-source benchmark  comprising $650$k (architecture, accuracy) pairs evaluated on CIFAR-10. By unifying multiple search spaces into a single ONNX-based format, it enables training and evaluation of NAS surrogates across diverse architecture families.

    \item We introduce \textbf{ONNX-Net}, a universal neural architecture representation and surrogate modeling framework built on ONNX-Bench. ONNX-Net captures complete architectural structure, including topology, operator types, and fine-grained parameterisation, in a compositional and extensible representation that is independent of predefined search spaces.

    \item We show that an LLM-based surrogate using ONNX-Net achieves strong predictive performance and zero-shot transferability. Notably, our approach shows exceptional sample efficiency, matching concurrent approaches with less than 1\% of the training data.
\end{itemize}

\section{Related Work}

\paragraph{Network Search Spaces}{
The effectiveness of NAS depends strongly on the search space design.
Early work explored ResNet-style \citep{HeZRS16} building blocks, leading to the dominant cell-based designs---where a single cell structure is searched and stacked to form the network \citep{ying2019bench, dong2020bench, dong2021nats, liu2018darts, siems2020bench}. While computationally efficient, cell-based spaces are very restrictive, motivating more expressive hierarchical and
grammar-based search spaces~\citep{schrodi2023construction, ericsson2024einspace} offer a more flexible and principled way of generating diverse architectures with different macro and micro structures.

Despite these advances, search spaces have been evaluated in isolation.
Recent efforts to unify benchmarks often use Python code as a representation \citep{Rahman2025lemonade, Gao2025llm4gnas, Zhou2025lapt, Nasir2024llmatic, Chen23evoprompting}. While very general, this representation has drawbacks. The same architecture can be implemented with markedly different code across different deep learning libraries. Even within a single framework, the same model can be expressed in many syntactically distinct ways, creating a many-to-one mapping from Python code to the underlying model.

In this paper, we take a more holistic view by introducing ONNX-Bench, a dataset that unifies multiple search spaces under a common ONNX format. This enables performance prediction and search methods to operate across diverse search spaces within a consistent representation, enabling fair comparison and more general NAS approaches. Another central advantage of structured file formats such as ONNX is the high expressivity, while yielding far fewer distinct encodings per model. We hypothesise that ONNX representations are substantially more sample-efficient than Python code representations, because the model need not learn invariances to a large variety of superficial code-level differences.}

\begin{table}[t]
    \centering
    \caption{Composition of the ONNX-Bench dataset. The benchmark aggregates $\sim$650k architectures across cell-based and hierarchical search spaces, with evaluations on CIFAR-10, UnseenNAS, and TransNAS-Bench-101 tasks.}
    \vspace{0.1em}
    \label{tab:onnx_bench}
    \resizebox{0.55\textwidth}{!}{%
    \begin{tabular}{lllr}
    \toprule
    Search Space              & Type                          & Evaluation                & Num Arch. \\
    \midrule
    NAS-Bench-101             & \multirow{5}{*}{Cell-based}   & \multirow{4}{*}{CIFAR-10} & 423 624            \\
    NAS-Bench-201             &                               &                           & 15 625             \\
    NATS-Bench                &                               &                           & 32 768             \\
    NAS-Bench-301             &                               &                           & 57 189             \\ \cmidrule(lr){3-3}
    TransNAS-Bench-101        &                               & Other                     & 38 895             \\ \cmidrule(lr){1-1}\cmidrule(lr){2-2}\cmidrule(lr){3-3}\cmidrule(lr){4-4} \addlinespace[0.5ex]
    hNAS-Bench-201            & \multirow{3}{*}{Hierarchical} & \multirow{2}{*}{CIFAR-10} & 8 000              \\ \cmidrule(lr){1-1}
    \multirow{2}{*}{einspace} &                               &                           & 57 495             \\ \cmidrule(lr){3-3}
                              &                               & UnseenNAS                 & 16 000             \\
    \midrule
    Total                     &                               &                           & 649 596            \\ \bottomrule
    \end{tabular}%
    }
\end{table}

\noindent\textbf{Network Encodings}{
Orthogonal to search space design, the network encoding plays a crucial role for both the search and performance prediction methods. Cell-based search spaces typically use adjacency matrices with Graph Neural Networks (GNNs) \citep{arch2vec20, Ning_DGF, Velickovic_GAT}, with recent extensions incorporating information flow \cite{hwang2024flowerformer, kim2025learning}, or causal features \cite{ji2025carl}.
To overcome the pure adjacency-based structure, \cite{LukasikMK25} use zero-cost proxies as architecture encodings, and \cite{KadlecovaLPVSNH24} add topology information as tabular features. Cross-space transfer has also been explored via contrastive graph encoders \cite{mills2023gennape} and combinations of search space-specific learned encodings \cite{akhauri2024encodings}.
However, these network representations are search space-specific, and empirical validation remains confined to cell-based search spaces, limiting flexibility and scalability. Most related to our work, RLM \citep{akhauri2025regression} employs language models on raw ONNX text. However, the unprocessed representation introduces substantial noise and redundancy. We address this with a compact and informative encoding scheme that explicitly captures operation parameters and output shapes.}

\noindent\textbf{LLMs in NAS}{
LLMs have become increasingly used in NAS as performance predictors \citep{Jawahar2024llmpp} or architecture generation. Most approaches apply LLMs in settings where topology or operations are confined by some prior structure (e.g. a cell-based search space \citep{Cai2025seki, Zhong2024llmo, Zheng2023genius}, supernets \citep{Ji2025rznas, Jawahar2024llmpp}, choosing parameters only in predefined ranges \citep{Qin2024flnas}, or manipulating sequences, where entries have predefined meanings \citep{Dong2023hgnas, Hu2025lmsearcher}). \cite{qin2025transferrable} uses grammar-based string representation, with an LLM for performance prediction, outperforming zero-cost proxy and topology-based baselines.
However, all these methods are search space dependent and cannot be transferred from one search space to another. In this work, we aim to push the boundaries of NAS and present ONNX-Net, a universal network encoding, independent of the search space, capable of instant performance prediction. This encoding is independent of the search space design and allows encoding any neural network that can be converted into an ONNX file.}

\section{ONNX-Bench}
We introduce a new benchmark dataset for NAS and performance prediction. It collates networks across several sources in a unified format and evaluation setup. This is needed to take NAS beyond the restriction of individual smaller search spaces, and will enable the training of performance predictors that can transfer across existing and future search spaces.

To achieve high compatibility with different frameworks and network formats, we chose the ONNX file format \citep{onnx} as a basis for our work, as it is a de facto standard for neural network persistence. This allows our method, cf. \cref{sec:onnx_net}, to be applied to nearly any network found in the wild, as most frameworks and file formats support saving or conversion into ONNX.

ONNX is a binary file format that represents neural network architectures as directed graphs, where nodes are instances of a set of pre-defined operations, while edges represent tensors/arrays passed between these operations. Nodes also contain the hyperparameter values for their operations, e.g. the kernel size of a pooling operation. Additionally, every ONNX file contains a list of input tensors (and their values, in the case of learned parameters), and output tensors.

\begin{figure}
     \centering
     \begin{subfigure}[b]{0.58\textwidth}
         \centering
         \includegraphics[width=\textwidth]{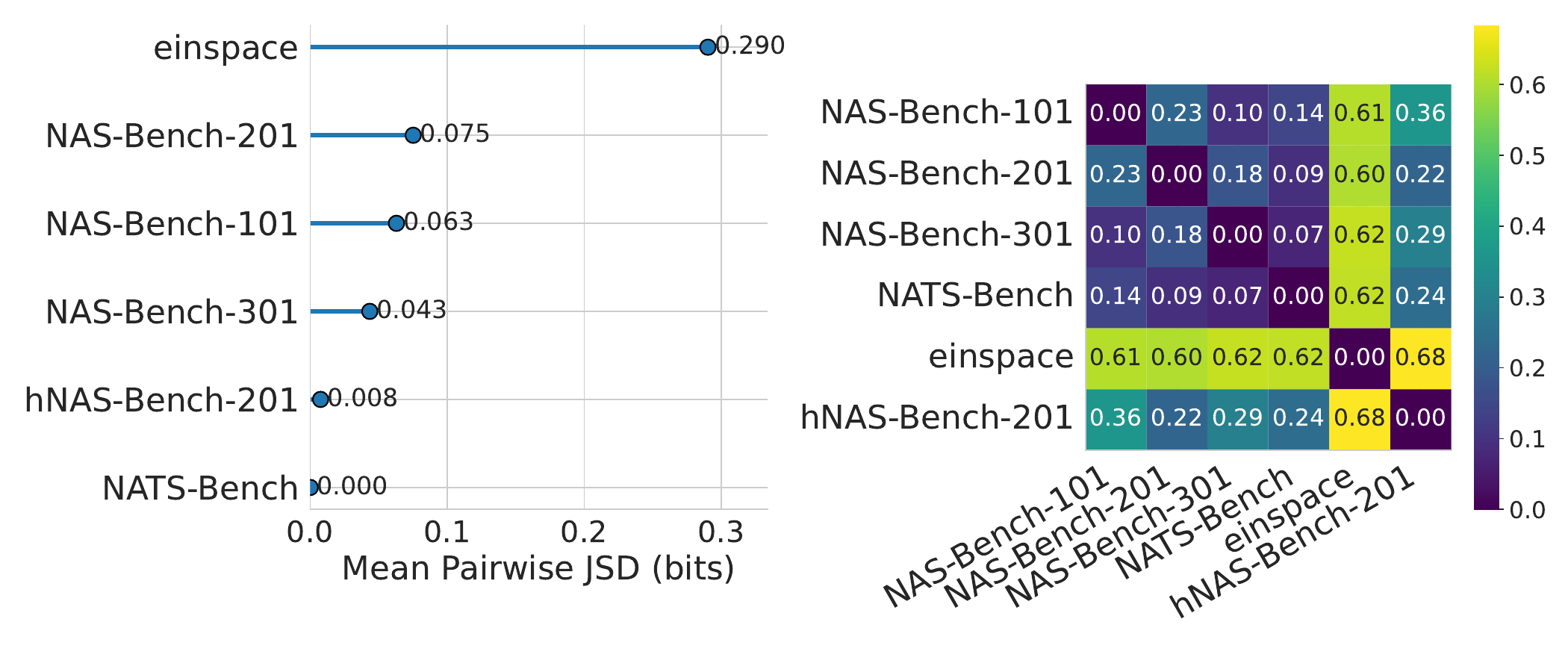}
         \caption{Diversity within (left) and between the search spaces (right) in ONNX-Bench, diversity is measured using Jensen-Shannon divergence (in bits).}
         \label{fig:space_diversity}
     \end{subfigure}
     \hfill
     \begin{subfigure}[b]{0.38\textwidth}
         \centering
         \includegraphics[width=\textwidth]{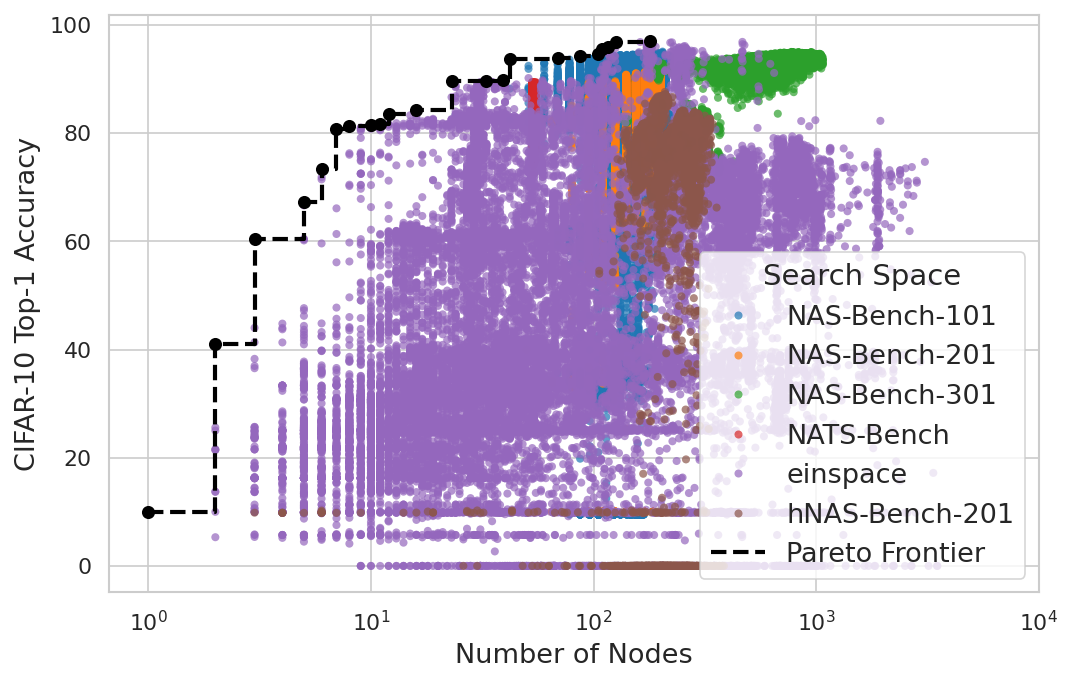}
         \caption{Distribution of nodes and CIFAR-10 accuracy for search spaces in ONNX-Bench with the Pareto frontier.}
         \label{fig:space_stat}
     \end{subfigure}
        \caption{}
        \label{fig:three graphs}
\end{figure}

The benchmark includes networks from multiple sources, spanning both cell-based and macro-level search spaces such as NAS-Bench-101 \citep{ying2019bench}, NAS-Bench-201 \citep{dong2020bench}, NATS-Bench  \citep{dong2021nats}, DARTS-style cells in NAS-Bench-302 \citep{siems2020bench}, and the hierarchical spaces hNAS-Bench-201 \citep{schrodi2023construction}, and einspace \citep{ericsson2024einspace}. Standardising these diverse architectures into ONNX allows them to be easily compared, reused, and extended. Table~\ref{tab:onnx_bench} summarises the distribution of architectures across sources, along with statistics such as the number of nodes and operator types.

All architectures are evaluated on the CIFAR-10 dataset using a consistent training pipeline. CIFAR-10 is widely used in NAS research and provides a balance between computational tractability and benchmark relevance. By fixing the dataset and training settings, ONNX-Bench ensures that observed performance differences reflect the architectures themselves rather than inconsistencies in training protocols.

In total, ONNX-Bench comprises 649 596 trained models, with node counts ranging from 1 to 3 503 and CIFAR-10 accuracies spanning [0.0, 97.03]. This diversity includes both poorly performing and competitive architectures, making the benchmark suitable for evaluating predictors across the full performance spectrum. Our ONNX-Bench aims to support the development of NAS methods that transfer across search spaces and reduce the need for repeated, costly retraining of architectures from scratch.

To further analyse the difference between each search space, we calculate diversity metrics both within and between search spaces in ONNX-Bench. For each model, we count ONNX \texttt{node.op\_type} occurrences (excluding Constant) and normalise to a probability $p$ over the op vocabulary $V$. For a sampled set $S$ of $n$ models from a search space. We compute Jensen–Shannon divergence (JSD, base-2, in bits) between all pairs:

\begin{equation}
\begin{aligned}
JSD(p_i, p_j) = \frac{1}{2} KL(p_i \,\|\, m) + \frac{1}{2} KL(p_j\,\|\,m), \quad \text{where } m = \frac{1}{2}(p_i + p_j)
\end{aligned}
\end{equation}

In terms of across-space dissimilarity, for two spaces $A$ and $B$, we pool $n$ sampled models in each space to obtain $p_{pool}^A$ and $p_{pool}^B$ over joint vocabulary $V=V_A \cup V_B$, then report $JSD(p_{pool}^A,p_{pool}^B)$ in bits.

The ideal NAS search space encompasses all well-performing neural network architectures possible. While this is infeasible to achieve, we argue that a very diverse search space comes closest to this goal. In \cref{fig:space_diversity}, we report diversity measures over 5k random samples from each search space, and show that hierarchical search spaces, such as einspace and hNAS-Bench-201, differ strongly from other spaces (and, in the case of einspace, also have a much wider variety of architectures). That these additional architectures are not only composed out of low-performing "fail-cases" can be clearly seen in \cref{fig:space_stat}. As ONNX-Bench encompasses all aforementioned search spaces, it fulfils the goal of architecture diversity to a high degree. 

\section{ONNX-Net}\label{sec:onnx_net}

As a baseline for future performance predictors developed on ONNX-Bench, we propose a presentation that allows describing any neural network architecture in the ONNX format and can be easily coupled with an instant performance prediction on a given dataset.

The central part of our proposed approach is the representation of the neural architecture in the form of natural language. To represent the information contained in the ONNX files as text, we first reconstruct the graph contained in the file in memory, retaining all information. 
Since context length is a limiting factor for many LLMs, we try to reduce the size of the graph by performing a variety of lossless optimisations described in \cref{sec:graph optimisations}.

The resulting, shortened graph is again shrunk by merging chains of operations without any branches, to obtain the final, condensed version of the graph.
We then convert this graph structure into text by printing each chain of nodes on one line, in the following format:

\vspace{0.4em}
\begin{minipage}{\linewidth}
\begin{verbatim}
    Op(Input1, ...)(Param1=Value,...) --> Op(prev, ...)(Param1=Value,...)
      --> ... --> Output1, ..., OutputN:Shape
\end{verbatim}
\end{minipage}
\vspace{0.1em}

\begin{table}[]
\centering
\caption{NAS-Bench-301 in domain performance in Kendall's $\tau$. We use 100 samples for training to match the setup from RLM.}
\label{tab:indomain_darts}
\resizebox{0.85\textwidth}{!}{%
    \begin{tabular}{lccccccc}
        \toprule
                  & MLP   & Arch2Vec & CATE  & GNN   & $FLAN^T$ & RLM   & ONNX-Net \\ \cmidrule(lr){1-1}\cmidrule(lr){2-8}
        NAS-Bench-301(DARTS) & 0.124 & 0.333    & 0.425 & 0.523 & \textbf{0.567}                   & 0.528 & \underline{0.546}   \\
        \bottomrule
    \end{tabular}
}
\end{table}

To show the applicability of ONNX-Net we train an LLM to act as an exemplary surrogate model. NAS is prohibitively expensive due to the need to train every candidate architecture to evaluate its performance. Surrogate models circumvent this recurring cost by acting as a performance predictor, inferring the performance of a candidate architecture of some search space (in this case, all architectures representable by ONNX) by information obtained without training the candidate. This accelerates the search, as the inference speed of many neural networks is negligible in comparison to the amount of time a (partial) training may require.

\newcommand{\predictor}{\ensuremath{P_\theta}}
\newcommand{\allarchs}{\mathcal{A}}
\newcommand{\dataset}{\mathcal{D}}
\newcommand{\trainfunc}{\phi}
\newcommand{\lossfunc}{\mathcal{L}}
Formally, our goal is to create a learned predictor $\predictor(\cdot): \allarchs \to \mathbb{R}$, which is capable of ranking any architecture in $\allarchs$ by its performance on a certain dataset in a certain metric, e.g. ranking architectures by their accuracy on CIFAR10. Here, $\allarchs$ denotes the space of all possible ONNX-encoded neural network architectures. We further define $\trainfunc(\cdot \,; \dataset): \allarchs \to \mathbb{R}$ as the training process for any network $a \in \allarchs$ on dataset $\dataset$, which returns the validation performance. Evaluating $\trainfunc$ is computationally expensive, therefore, minimizing the number of evaluations is the primary motivation for creating a surrogate predictor.\\
The parameters $\theta$ of $\predictor$ are trained on samples of $\{$architecture, accuracy$\}$  pairs $\{(a_i, \trainfunc(a_i; \dataset))\}_{i=1}^N$ by minimizing an empirical risk objective 
\begin{equation}
    \hat{\theta} = \arg\min_{\theta} \frac{1}{N} \sum_{i=1}^N 
    \lossfunc\left(\predictor(a_i), \, \trainfunc(a_i; \dataset)\right).
    \label{eq:f_optimisation}
\end{equation}
$\mathcal{L}$ is a loss function that quantifies the discrepancy between the predicted and true performances. This loss can take the form of e.g. a mean-squared error or a ranking loss.

\section{Experiments}
ONNX-Bench can be a valuable dataset for the NAS community to investigate search space-independent surrogate models that generalise to a diverse range of architectures. As a baseline, we evaluate ONNX-Net with respect to its ability to predict performance on new search spaces (\cref{sec:loo}), with respect to its zero-shot performance (\cref{sec:zero-shot}), and its ability to generalise to new tasks (\cref{sec:unseen_task}).

\begin{figure}
     \centering
     \begin{subfigure}[b]{0.48\textwidth}
         \centering
         \includegraphics[width=\textwidth]{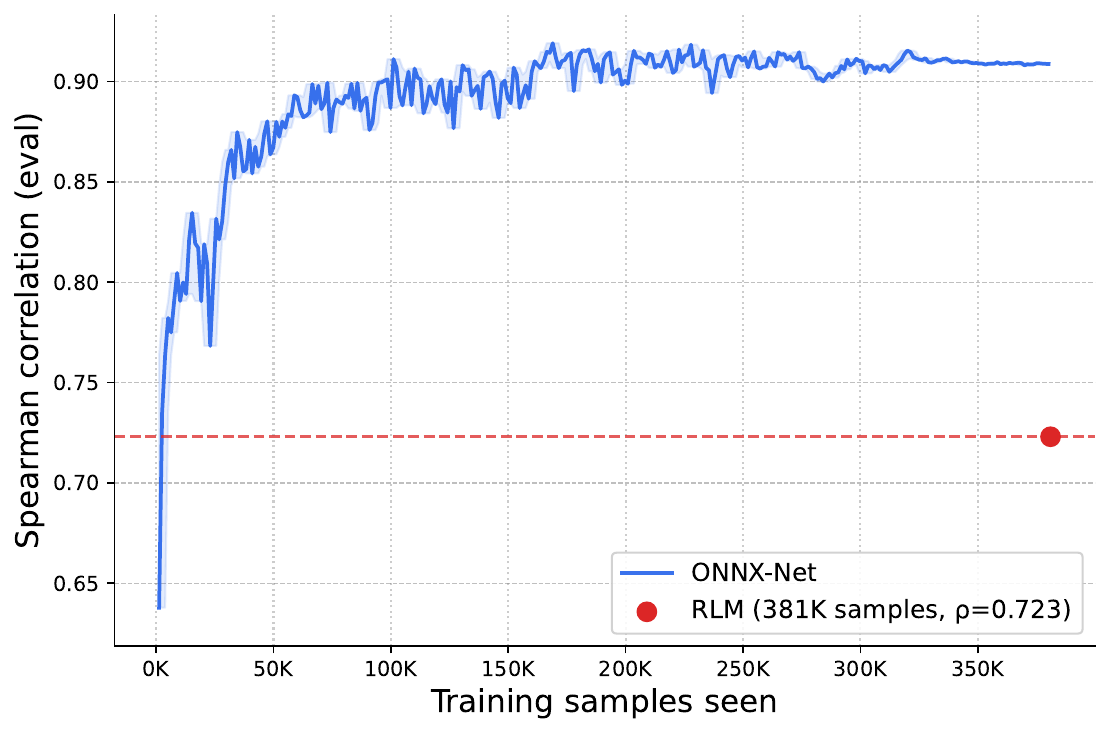}
         \caption{Spearman correlation against number of training samples mirroring RLM settings. ONNX-Net is drastically more sample efficient, reaching the same correlation with 150$\times$ less data.}
         \label{fig:pretrain_scratch}
     \end{subfigure}
     \hfill
     \begin{subfigure}[b]{0.48\textwidth}
         \centering
         \includegraphics[width=\textwidth]{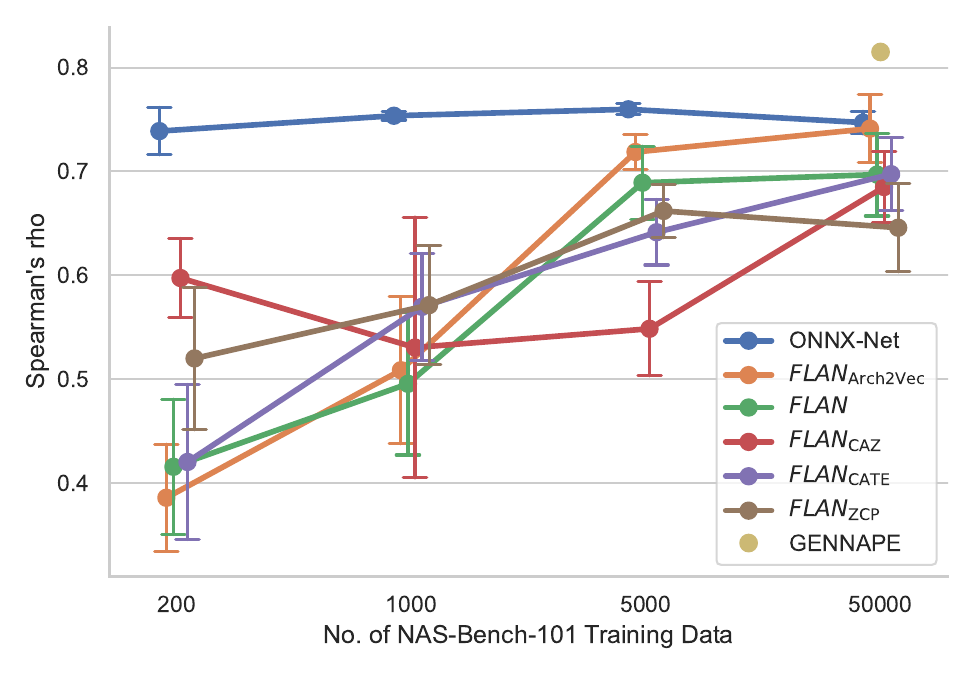}
         \caption{Zero-shot predictor trained on different number of \texttt{NAS-Bench-101}, evaluated on NAS-Bench-201. Avg. Spearman's $\rho$ and standard error over 5 random seeds is reported.}
         \label{fig:zero-shot}
     \end{subfigure}
        \caption{}
        \label{fig:three graphs 2}
\end{figure}

\noindent\textbf{Metrics}
We report the rank correlation values (Kendall’s $\tau$, Spearman's $\rho$) between the ground-truth and predicted performances.

\noindent\textbf{Model}
We fine-tune a \texttt{ModernBERT-large} model as the performance predictor, and we also compare it with other LLMs in \cref{sec:base model}.

\subsection{In-Domain Performance and Baseline Comparisons} \label{sec:indomain}
We first evaluate ONNX-Net in an in-domain setting to establish its ceiling performance relative to specialized and general surrogates \citep{akhauri2025regression, qin2025transferrable, akhauri2024encodings}.

\noindent\textbf{Comparison with General-Purpose Surrogates}
We evaluate on NAS-Bench-301 using 100 training samples to match the setting in \cite{akhauri2025regression}. As shown in \cref{tab:indomain_darts}, ONNX-Net outperforms RLM and GNN-based predictors. While FLAN achieves a slightly higher correlation, it relies on significant external information such as Zero-Cost Proxies and custom pre-processing, whereas ONNX-Net operates purely on the architecture's text representation.

\noindent\textbf{Sample Efficiency and Pre-training}
We also compare training from scratch using the RLM protocol (4-layer LM pretrained on 381k NAS-Bench-101 samples). Results in \cref{fig:pretrain_scratch} show that ONNX-Net reaches a Spearman's $\rho$ of 0.919 compared to RLM's 0.723. Notably, ONNX-Net achieves the same performance level as the fully trained RLM after seeing only 2.5k samples (150$\times$ better sample efficiency). We attribute this superiority to our optimized text-based ONNX encoding. While RLM uses a representation closer to raw ONNX containing redundant information, our encoding condenses the graph and enriches it with specific input parameters, which enables substantially higher sample efficiency.

\subsection{How well does the surrogate perform on new search spaces?} \label{sec:loo}
We assess how well the surrogate performs when trained on all but one search space, i.e.~how it generalises to the excluded search space (\cref{tab:leave_one_out}). For better comparison, we also include the surrogate performance when trained on all search spaces. Given that NATS-Bench is an extension of NAS-Bench-201, they are considered together as NATS-Bench in this experiment. The detailed train/val split regime is described in \cref{sec:split}.

Results in \cref{tab:leave_one_out} show that training on all search spaces yields strong but not uniformly optimal Kendall’s $\tau$ across targets. Holding out a search space typically degrades performance on that space, with a notable exception for hNAS-Bench-201, where leaving it out actually improves transfer (0.533 → 0.565), suggesting negative transfer when hNAS is included. The best per-column scores are generally achieved without the full mixture, suggesting future work regarding finding the optimal data mixture for a universal surrogate model.

\begin{table*}[t]
\centering
\captionsetup{skip=2pt}
\caption{Kendall's $\tau$ correlations for surrogate models trained on the full set of search spaces as well as all-but-one search space. Rows after the first show the spaces left out of the training set, and columns show the evaluation spaces where we compute correlations on held-out data.}
\label{tab:leave_one_out}
\resizebox{0.85\textwidth}{!}{%
\begin{tabular}{clccccc}
\toprule
               {} & & NAS-Bench-101 & NATS-Bench & NAS-Bench-301 & hNAS-Bench-201 & einspace  \\ \cmidrule(lr){1-2}\cmidrule(lr){3-7} \addlinespace[0.5ex]
{} & Train on all     & 0.772         & 0.788      & 0.691         & \underline{0.533}          & 0.477     \\ \cmidrule(lr){1-2}\cmidrule(lr){3-7} \addlinespace[0.5ex]
\multirow{5}{*}{\rotatebox[origin=c]{90}{Leave-out}} & NAS-Bench-101  & 0.529         & 0.815      & 0.687         & 0.386          & \textbf{0.529}     \\ 
{} & NATS-Bench     & 0.744         & 0.390      & \textbf{0.704}         & 0.382          & 0.478     \\ 
{} & NAS-Bench-301  & 0.777         & 0.819      & 0.508         & 0.214          & 0.474     \\
{} & hNAS-Bench-201 & \underline{0.787}         & \underline{0.825}      & 0.693         & \textbf{0.565}          & \underline{0.524}     \\
{} & einspace       & \textbf{0.794}         & \textbf{0.843}      & \underline{0.694}         & 0.456          & 0.301     \\ \bottomrule
\end{tabular}%
}
\end{table*}

\subsection{Zero-shot performance across search spaces} \label{sec:zero-shot}

To enable comparison with prior work~\citep{akhauri2024encodings, mills2023gennape}, we evaluate the zero-shot transfer from \texttt{NAS-Bench-101} to \texttt{NAS-Bench-201}. Concretely, we train the surrogate on 50k random  $\{$architecture, accuracy$\}$  pairs from \texttt{NAS-Bench-101} and evaluate on the full \texttt{NAS-Bench-201} set without any adaptation.

We also study data scaling by varying the \texttt{NAS-Bench-101} training set size: 200, 1k, and 5k samples. We replicate the FLAN setup using their released code under our protocol; we are unable to replicate GENNAPE due to the lack of reproducible code, and therefore we report only results from their paper.

\begin{table}[t]
\centering
\footnotesize
\captionsetup{skip=2pt}
\caption{Zero-shot predictor trained on 50k samples from NAS-Bench-101 and evaluated on NAS-Bench-201. Avg. Spearman's $\rho$ over 5 random seeds is reported.}
\label{tab:zero-shot}
\begin{tabular}{cccccccc}
\toprule
\multirow{2}{*}{Transfer} & \multirow{2}{*}{GENNAPE} & \multicolumn{5}{c}{FLAN}                 & \multirow{2}{*}{ONNX-Net} \\
                          &                          & -     & CATE  & Arch2Vec & ZCP   & CAZ   &     \\ \cmidrule(lr){1-1}\cmidrule(lr){2-2}\cmidrule(lr){3-7}\cmidrule(lr){8-8} \addlinespace[0.5ex]
Zero-Shot                 & \textbf{0.815}           & 0.697 & 0.697 & 0.741    & 0.646 & 0.685 & \underline{0.747} \\ \bottomrule              
\end{tabular}
\end{table}

Results in \cref{tab:zero-shot} and \cref{fig:zero-shot} show that GENNAPE achieves the strongest zero-shot transfer overall ($\rho{=}0.815$), noting that it utilises an ensemble combining multiple predictors with two pairwise classifiers. Relative to FLAN, our surrogate consistently achieves higher zero-shot performance across all training-set sizes, including FLAN variants that incorporate additional encodings such as CATE, Arch2Vec, or zero-cost proxies (ZCP); the gains are largest in the low-data regime. Our surrogate reaches its peak zero-shot correlation with 5k training samples and exhibits substantially lower seed-to-seed variance than FLAN.

According to the analysis in \cref{fig:space_diversity}, the Jensen-Shannon divergence between NAS-Bench-101 and NAS-Bench-201 is 0.23, which explains the good zero-shot transfer ability. Additionally, we also explore zero-shot transfer ability across more distinct search spaces. Results in \cref{tab:zero-shot einspace} show weaker transfer performance when divergence is high. This outcome underscores our motivation for ONNX-Bench: gathering a unified, diverse architecture collection to expose the surrogate to heterogeneous architectural patterns and improve generalisation to truly unseen spaces.

\subsection{How well does the surrogate do on new datasets?}\label{sec:unseen_task}
We also assess the ability of the surrogate model to generalise to classification tasks beyond CIFAR10.

\noindent\textbf{CIFAR-10 to Unseen NAS on einspace (cross-dataset).}
We train a surrogate using CIFAR-10 labels and evaluate zero-shot on eight Unseen NAS \citep{geada2024insights} datasets within the einspace search space. Results in \cref{tab:unseennas} show that including einspace during training (\emph{Full}) markedly improves zero-shot transfer to UnseenNAS. \emph{Full} outperforms \emph{w/o einspace} on 6 of 8 datasets. Overall, including einspace training data is critical for cross-dataset generalisation within the einspace search space, though the optimal training mix may be task-dependent.

\begin{table*}[t]
\centering
\caption{Kendall's $\tau$ correlations for zero-shot transfer from CIFAR-10 to Unseen NAS datasets with or without search space-specific data.}
\label{tab:unseennas}
\resizebox{0.9\textwidth}{!}{%
\begin{tabular}{@{}ccccccccc@{}}
\toprule
             & AddNIST & Language & MultNIST & CIFARTile & Gutenberg & Isabella & GeoClassing & Chesseract \\ \cmidrule(lr){1-1}\cmidrule(lr){2-9} \addlinespace[0.5ex]
Full         & 0.364   & 0.338    & 0.449    & 0.351     & 0.582     & 0.248    & 0.095       & 0.294    \\ \cmidrule(lr){1-1}\cmidrule(lr){2-9} \addlinespace[0.5ex]
w/o einspace & 0.156   & 0.131    & 0.245    & 0.058     & 0.317     & 0.135    & 0.249       & 0.302    \\ \bottomrule
\end{tabular}%
}
\end{table*}

 \begin{table}[]
\footnotesize
\centering
\caption{Zero-shot predictor trained on CIFAR-10 labels and evaluated on CIFAR-100 and ImageNet16-120. Avg. Kendall's $\tau$ over 5 random seeds is reported.\label{tab:cifar100}}
\begin{tabular}{lccc}
\toprule
                              & \multicolumn{3}{c}{Train Size}                                                                                      \\
\multirow{-2}{*}{Target Task} & 200                                                  & 1000                                                 & 5000  \\ \cmidrule(lr){1-1}\cmidrule(lr){2-4} 
CIFAR-100                     & 0.431                                                & 0.499                                                & 0.609 \\
ImageNet16-120                &  0.419 &  0.480 & 0.593 \\ \bottomrule
\end{tabular}
\end{table}

\noindent\textbf{CIFAR-10 to CIFAR-100 and ImageNet16-120 on NAS-Bench-201.}
Additionally, we train a surrogate using different sample sizes of NAS-Bench-201 architectures using their respective CIFAR-10 labels and evaluate its cross-task ability on the evaluation set in a zero-shot setting. Results in \cref{tab:cifar100} show that the surrogate successfully transfers architectural knowledge across distinct classification tasks.

\subsection{Ablation Study}
We decompose the architecture-to-text encoding into four components (\cref{fig:encoding ablation}):
(i) \emph{Base information}: operation name and output index;
(ii) \emph{Input information}: weight/bias shapes and names of the inputs to each operation;
(iii) \emph{Parameter information}: operation-specific parameters (e.g., kernel shape and padding for convolutions);
(iv) \emph{Output shape information}: the tensor shape of operation’s output.
To assess the importance of each component, we compare:
\begin{figure}[t]
    \centering
    \includegraphics[width=0.75\linewidth]{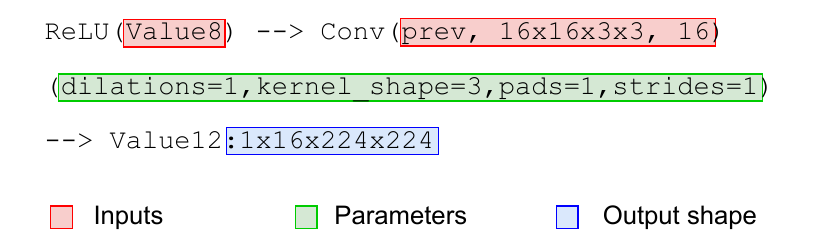}
    \caption{An excerpt of our text encoding for some neural ar. Highlighted are the different components of the encoding that we evaluate during the ablation study.}
    \label{fig:encoding ablation}
\end{figure}

(a) a \emph{base} variant using only the Base information;
(b) the \emph{full} encoding (all four components);
(c) three variants where we add one component to the base variant.
All models are trained and evaluated under the same setting as \cref{sec:zero-shot}, we additionally add zero-shot experiments from NAS-Bench-201 to NAS-Bench-101, excluding the 50k variant due to data limitations for NAS-Bench-201.

\begin{table*}[t]
    
\centering
\caption{Zero-shot transfer learning between NAS-Bench-101 and NAS-Bench-201 using different text encodings. Avg. Spearman's $\rho$ over 5 random seeds is reported. We also report Avg. Kendall's $\tau$ in \cref{tab:encoding ablation kendall}.}
\label{tab:encoding ablation}
\resizebox{0.55\textwidth}{!}{%
\begin{tabular}{lccccccc}
\toprule
\multirow{2}{*}{Encoding} & \multicolumn{4}{c}{NB101 → NB201}                         & \multicolumn{3}{c}{NB201 → NB101} \\ 
                          & 200            & 1000           & 5000           & 50000          & 200          & 1000         & 5000        \\ \cmidrule(lr){1-1}\cmidrule(lr){2-5}\cmidrule(lr){6-8} \addlinespace[0.5ex]
Base                      & 0.618          & 0.644          & 0.691          & 0.666          & 0.599        & 0.667        & 0.691       \\ \cmidrule(lr){1-1}\cmidrule(lr){2-5}\cmidrule(lr){6-8} \addlinespace[0.5ex]
+Inputs                   & \textbf{0.746} & \underline{0.752}          & \underline{0.756}          & \underline{0.743}          & 0.682        & \textbf{0.755}        & \underline{0.780}       \\
+Parameters               & 0.726          & 0.724          & 0.715          & 0.718          & \underline{0.713}        & 0.720        & 0.762       \\
+Out Shape                & 0.660          & 0.679          & 0.693          & 0.682          & 0.658        & 0.686        & 0.675       \\ \cmidrule(lr){1-1}\cmidrule(lr){2-5}\cmidrule(lr){6-8} \addlinespace[0.5ex]
Full                      & \underline{0.739}          & \textbf{0.754} & \textbf{0.760} & \textbf{0.747} & \textbf{0.715}        & \underline{0.740}        & \textbf{0.781}       \\ \bottomrule
\end{tabular}%
}
\end{table*}

Table~\ref{tab:encoding ablation} shows that enriching the encoding with \emph{Input information} yields the largest single-step gain over the \emph{Base} variant across both settings, highlighting the importance of explicit connectivity and weight shape cues at the inputs. Adding only \emph{Parameter information} helps when data size is small but offers diminishing gains as the train size grows. \emph{Output shape} alone provides only marginal improvements over \emph{Base}, which matches the expectation as it adds the least amount of information. The \emph{Full} version lags slightly behind +Inputs for some cases, likely due to the longer sequences reducing sample efficiency, but best overall. We also observe a mild dip at 50k relative to 5k for most variants, consistent with the observation in \cref{sec:zero-shot}; this points to potential overfitting to the source domain.

To study the effect of the different components of ONNX-Net, we conduct a series of ablation studies. Ablations on the loss function, base model, and tokenizer are detailed in \cref{sec:add_ablation}.

\section{Limitations}
While our proposed method demonstrates robust performance across the evaluated benchmarks, we acknowledge several inherent limitations that warrant future investigation. First, ONNX-Bench relies on a compilation of pre-existing NAS datasets. Consequently, this source material inherently restricts the distribution of architectural designs and tasks other than low-resolution image classification. Secondly, direct comparison of our final performance using ONNX-Net with existing baseline approaches presents notable methodological challenges. This difficulty stems from the novel architecture and the generality of our methodology, which prevents the straightforward application of established approaches with identical settings. Finally, although ONNX-Net achieves exceptional results within the zero-shot and few-shot learning paradigms, its performance scalability suggests that further improvements can be made regarding the usage of larger volumes of training data.

\section{Conclusion}
We have introduced ONNX-Bench, a unified collection of NAS benchmarks containing architectures in a shared ONNX format and performance scores on a common dataset, CIFAR-10. This benchmark can be used to evaluate performance predictors on a general suite of architectural styles, going beyond the existing narrow cell-based benchmarks. In the future, we hope to use it for developing general search methods as well.

Using this novel benchmark, we develop and evaluate a novel surrogate model we call ONNX-Net. It uses our condensed string encoding of the ONNX representation as the input to an LLM, fine-tuned towards the performance prediction task. It shows very strong zero-shot performance using only a small amount of training data. Compared to previous methods, it can handle more general and flexible architecture inputs, though it also becomes clear that the problem is more difficult and that more restricted graph-based approaches can outperform our more generally applicable method. We hope that the release of this benchmark, encoding, and surrogate can spur more research into search space-agnostic NAS.

Future work can build upon the benchmark to create search methods in a general architecture format such as ONNX. Continuing in the direction of LLM and string representations can lead to guided generation of architecture candidates. Furthermore, we acknowledge the limitation of this work to mainly focus on performance on the CIFAR-10 dataset, and hope that future work will expand the capabilities of surrogates to take the dataset as context. Finally, due to its general scope, we hope to continue expanding ONNX-Bench as a living benchmark, to make it even more diverse, e.g.~with attention-based architectures.

\section*{Acknowledgments}
This work was supported by the Engineering and Physical Sciences Research Council [EP/X020703/1], a studentship from the School of Engineering at the University of Edinburgh. Jovita Lukasik acknowledges support by the Deutsche Forschungsgemeinschaft (German Research Foundation, DFG) under the project number 572735628.

\bibliography{bibliography}




\newpage
\appendix
\noindent{\large \bfseries Appendix\\}

\noindent The appendix contains supplementary methodological and experimental details:
\begin{itemize}
    \item Appendix~\ref{sec:broader impact} discusses broader impacts,
    \item Appendix~\ref{sec:graph optimisations} describes the graph optimisations used to construct compact architecture representations,
    \item  Appendices~\ref{sec:additional experiments} and~\ref{sec:add_ablation} provide additional experiments and ablations, including comparisons with search space-specific methods and analyses of the loss function, model backbone, and tokenizer,
    \item Appendices~\ref{sec:unseen nas} and~\ref{sec:split} describe the Unseen NAS datasets and the data-splitting protocols,
    \item  Appendix~\ref{sec: search results} reports supplementary architecture-search results,
    \item Appendix~\ref{sec:implementation} provides implementation and computational details,
    \item Appendices~\ref{sec:graph optim} and~\ref{sec:string encoding} present additional graph-optimisation visualisations and string-encoding examples.
\end{itemize}

\section{Broader Impact} \label{sec:broader impact}

Fine-tuning language models is a relatively computationally heavy task, which has a potential negative effect on the environment. However, our framework lowers the computational barriers to high-performance architecture search, enabling researchers with limited hardware resources to discover optimized models without the prohibitive costs of training thousands of candidates from scratch.

\section{Graph Optimisations} \label{sec:graph optimisations}

\begin{description}
    \item[Node removal] We remove nodes we deem to be of low importance, such as identity operations or input nodes for parameters, which can be implicitly inferred by their context.
    \item[Subgraph merging] We merge common subgraphs with known interpretation into single nodes, e.g. merging a matrix-vector multiplication with a parameter, followed by an addition with a parameter, into a single node, corresponding to a linear layer.
\end{description}
Some of the optimisations were performed by the \textit{ONNX-Simplifier} tool \citep{onnxsim}.

\section{Additional Experiments} \label{sec:additional experiments}

\paragraph{Comparison with Search Space Specific Methods}
Finally, we compare ONNX-Net against space-specific baselines on the \textit{einspace} search space. As shown in \cref{tab:einspace_specific}, ONNX-Net achieves competitive performance (Kendall’s $\tau=0.513$) matching traditional surrogates. While an LM using domain-specific grammar \citep{qin2025transferrable} holds an edge in its native environment, ONNX-Net’s ability to match traditional surrogates without any search space-specific priors validates the flexibility of our universal representation.

\begin{table}[htbp]
\centering
\caption{einspace in-domain performance mirroring \citep{qin2025transferrable} settings. Kendall's $\tau$ is reported.}
\label{tab:einspace_specific}
\resizebox{0.55\textwidth}{!}{%
\begin{tabular}{lcccc}
\toprule
 & RF    & XGB   & Qin et. al.\cite{qin2025transferrable} & ONNX-Net \\ \cmidrule(lr){1-1}\cmidrule(lr){2-5}
einspace   & 0.514 & 0.449 & 0.612 & 0.513   \\
\bottomrule
\end{tabular}
}
\end{table}

\begin{table}[t]
\centering
\caption{Zero-shot transfer learning among einspace, NAS-Bench-101 and NAS-Bench-201. Avg. Spearman's $\rho$ over 5 random seeds is reported. We also report Avg. Kendall's $\tau$ in \cref{tab:zero-shot einspace kendall}.}
\label{tab:zero-shot einspace}
\begin{tabular}{cccc}
\toprule
\multirow{2}{*}{Source → Target Search Space} & \multicolumn{3}{c}{Train Size} \\
           & 200         & 1000        & 5000        \\ \cmidrule(lr){1-1}\cmidrule(lr){2-4}
NAS-Bench-101 → einspace & 0.264 & 0.199 & 0.155 \\
einspace → NAS-Bench-101 & 0.310 & 0.351 & 0.348 \\
einspace → NAS-Bench-201 & 0.242 & 0.198 & 0.180 \\ \bottomrule
\end{tabular}
\end{table}

\section{Additional Ablations} \label{sec:add_ablation}

\subsection{Loss Function Choice}

While our framework treats NAS as a text-to-number task, the choice of loss function is critical for how the model interprets performance values. We justify our choice of Pairwise Ranking Loss by comparing it against standard regression objectives used in previous literature: Mean Squared Error (MSE) \citep{qin2025transferrable, lukasik2025better, fernandes2023devil} and Huber loss \citep{su2026llm}. We evaluate these using the ''Train on all'' setting across five search spaces.

\begin{table}[]
\centering
\caption{Loss function ablation mirroring the 'Train on all' setting in \cref{tab:leave_one_out}.}
\label{tab:loss ablation}
\resizebox{0.85\textwidth}{!}{%
\begin{tabular}{lccccc}
\toprule
Loss fn        & NAS-Bench-101  & NATS-Bench     & NAS-Bench-301  & hNAS-Bench-201                                       & einspace                                                      \\\cmidrule(lr){1-1}\cmidrule(lr){2-6} \addlinespace[0.5ex]
MSE              & 0.697          & 0.779          & 0.629          & \textbf{0.195}                                       & \textbf{0.528} \\
Huber            & 0.746          & \textbf{0.794} & 0.642          & 0.292                                                & \textbf{0.528}                                                \\
Pairwise Ranking & \textbf{0.772} & 0.788          & \textbf{0.691} & \textbf{0.533}                                       & 0.477  \\                                                      
\bottomrule
\end{tabular}
}
\end{table}

As shown in \cref{tab:loss ablation}, the Pairwise Ranking Loss consistently yields the highest Kendall correlation across most search spaces, particularly in complex domains like NAS-Bench-301 and hNAS-Bench-201. While regression-based losses like MSE and Huber attempt to predict absolute accuracy, our results suggest that the NAS surrogate benefits more from learning relative orderings. This confirms our hypothesis that for downstream architecture search, identifying the relative superiority of one candidate over another is more important than precise point estimation of accuracy.

\subsection{Base Model Choice} \label{sec:base model}

We evaluate the effect of the LM backbone by fine-tuning two families:
an encoder-based \texttt{ModernBERT} and a decoder-based \texttt{Qwen3}.
For each family, we use the same fine-tuning recipe, data, and evaluation protocol as in \cref{sec:zero-shot}, with results listed in \cref{tab:model ablation}.

\begin{table}[t]
\centering
\footnotesize
\caption{Zero-shot transfer learning from NAS-Bench-101 to NAS-Bench-201 using different LM backbone. Avg. Spearman's $\rho$ over 5 random seeds is reported. We also report Avg. Kendall's $\tau$ in \cref{tab:model ablation kendall}.}
\label{tab:model ablation}
\begin{tabular}{lcccc}
\toprule
\multirow{2}{*}{Model} & \multicolumn{4}{c}{NB101 → NB201} \\
                       & 200       & 1000       & 5000      & 50000      \\ \cmidrule(lr){1-1}\cmidrule(lr){2-5} \addlinespace[0.5ex]
ModernBERT-base                        & \underline{0.725}     & 0.730      & 0.744     &   0.737    \\
ModernBERT-large                         & \textbf{0.739}     & \textbf{0.754}      & \textbf{0.760}     &   \textbf{0.747}    \\ \cmidrule(lr){1-1}\cmidrule(lr){2-5} \addlinespace[0.5ex]
Qwen3-0.6b                        & 0.620     & 0.696      & \underline{0.747}     &   \underline{0.745}    \\
Qwen3-1.7b                       & 0.660     & \underline{0.735}      & 0.734     &   0.728    \\ \bottomrule
\end{tabular}
\end{table}

Across all data regimes, the encoder-based LM outperforms the decoder-based ones for zero-shot transfer from NAS-Bench-101 to NAS-Bench-201, with \texttt{ModernBERT-large} being consistently best. Scaling helps within the encoder family: \texttt{ModernBERT-large} surpasses \texttt{ModernBERT-base} at every data size. For \texttt{Qwen3}, the larger variant is better only when data size is small. The superiority of the encoder-based models matches the observation in \cite{qin2025transferrable}.

\subsection{Tokenizer Adaptation}

Given that our text representations can result in long input sequences, we investigate whether vocabulary adaptation can improve efficiency without sacrificing performance. We compare our original tokenizer against the vocabulary adaptation method proposed in \cite{nakash2025adaptivocab}, which compresses the text sequence by merging domain-specific tokens. We conduct this test on the \textit{einspace} search space, as its high structural diversity and long sequence lengths serve as a stress test for tokenization.

\begin{table}[]
\centering
\caption{Tokenizer ablation on einspace in-domain performance. Kendall's $\tau$ is reported}
\label{tab:tokenizer ablation}
\begin{tabular}{lc}
\toprule
Tokenizer       & einspace \\ \cmidrule(lr){1-1}\cmidrule(lr){2-2}
Original (Ours) & \textbf{0.513}    \\
AdaptiVocab     & 0.478   \\ \bottomrule
\end{tabular}
\end{table}

The results in \cref{tab:tokenizer ablation} indicate that while vocabulary adaptation successfully reduces average input lengths by approximately 50\%, it leads to a drop in Kendall correlation (0.513 vs. 0.478). This suggests that the information loss or the disruption of the pre-trained token embeddings during adaptation outweighs the computational benefits of shorter sequences. Consequently, we maintain the original tokenizer to preserve the full granularity of the architecture description.

\section{Unseen NAS Datasets} \label{sec:unseen nas}
In the following, we will provide an overview of the 8 Unseen NAS tasks \citep{geada2024insights} used in \cref{tab:unseennas}. All tasks are image classification tasks, comprising different difficulties. AddNIST is based on the MNIST dataset \citep{lecun-mnisthandwrittendigit-2010} using three channels laid on top of each other, with the label being the sum of the label of each channel. The language dataset aims to classify one of 10 languages. The language image is generated by randomly selecting four words from the target language, which are concatenated into a string. This string is then encoded into a $24 \times 24$ grid, where a black pixel indicates the presence of a letter at that grid position.
MultNNIST is similar to AddNIST but uses the multiplication of the three channels as a target. CIFARTile is a combination of four CIFAR-10 images in a $2\times2$ grid. The target here is the number of distinct CIFAR-10 classes shown in the grid.  Gutenberg aims at classifying authors based on three words of consecutive sequences encoded similarly to the Language dataset. The Isabella dataset classifies four music eras using recordings of these eras, which are converted into 64-band spectrograms. GeoClassing uses patches from the BigEarthNet \citep{SumbulCDM19} dataset to identify the corresponding country. Lastly, Chesseract contains images of the chess boards, of the final $15\%$ of the board state, of public games from eight grandmasters. The target here is to depict the classes: white wins, draw, black wins.

\section{Splitting method per space} \label{sec:split}
\begin{itemize}
    \item NAS-Bench-201 and NATS-Bench: Randomly sample 20\% of architectures as validation. Due to the high similarity between these two spaces, we merge them together.
    \item NAS-Bench-101: We adopt the validation indices provided by the paper \cite{akhauri2024encodings}.
    \item HNAS-Bench-201 and einspace: One search seed is reserved as validation split. For einspace, the validation mirrors the setup in \cite{qin2025transferrable}.
    \item NAS-Bench-301: Three sources are randomly picked as validation split.
\end{itemize}
Table~\ref{tab:data_stats} summarises the number of training and validation instances used from each search space.

\begin{table}[h]
\centering
\captionsetup{skip=2pt}
\caption{Dataset sizes per search space. For NAS-Bench-201 and NATS-Bench, we merge their respective train/validation partitions due to space similarity. All labels are CIFAR-10 top-1 accuracy.}
\begin{tabular}{lrr}
\toprule
Search space & Train & Validation \\
\midrule
NAS-Bench-101 & 40{,}000 & 7{,}290 \\
NAS-Bench-201 + NATS-Bench & 38{,}714 & 9{,}679 \\
NAS-Bench-301 & 40{,}000 & 5{,}892 \\
hNAS-Bench-201 & 6{,}403 & 1{,}000 \\
einspace & 37{,}416 & 1{,}582 \\
\bottomrule
\end{tabular}

\label{tab:data_stats}
\end{table}

\section{Search results} \label{sec: search results}

\begin{figure*}
    \centering
    \includegraphics[width=0.85\linewidth]{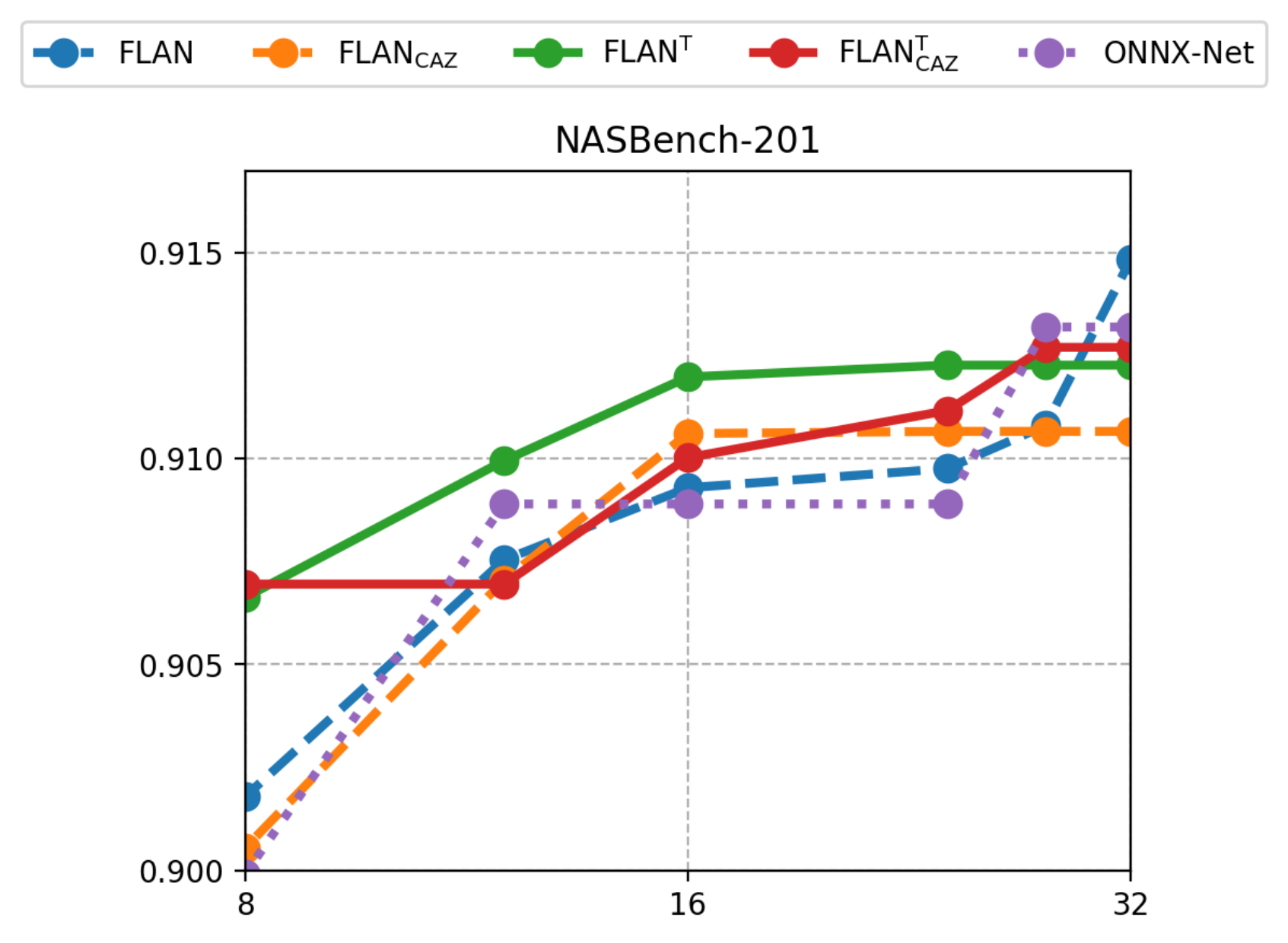}
    \caption{Comparison of the performance of different FLAN-Variants and our ONNX-Net approach using the FLAN-search on NASBench-201.}
    \label{fig:apdx_search_nb201}
\end{figure*}

\begin{figure*}
    \centering
    \includegraphics[width=0.85\linewidth]{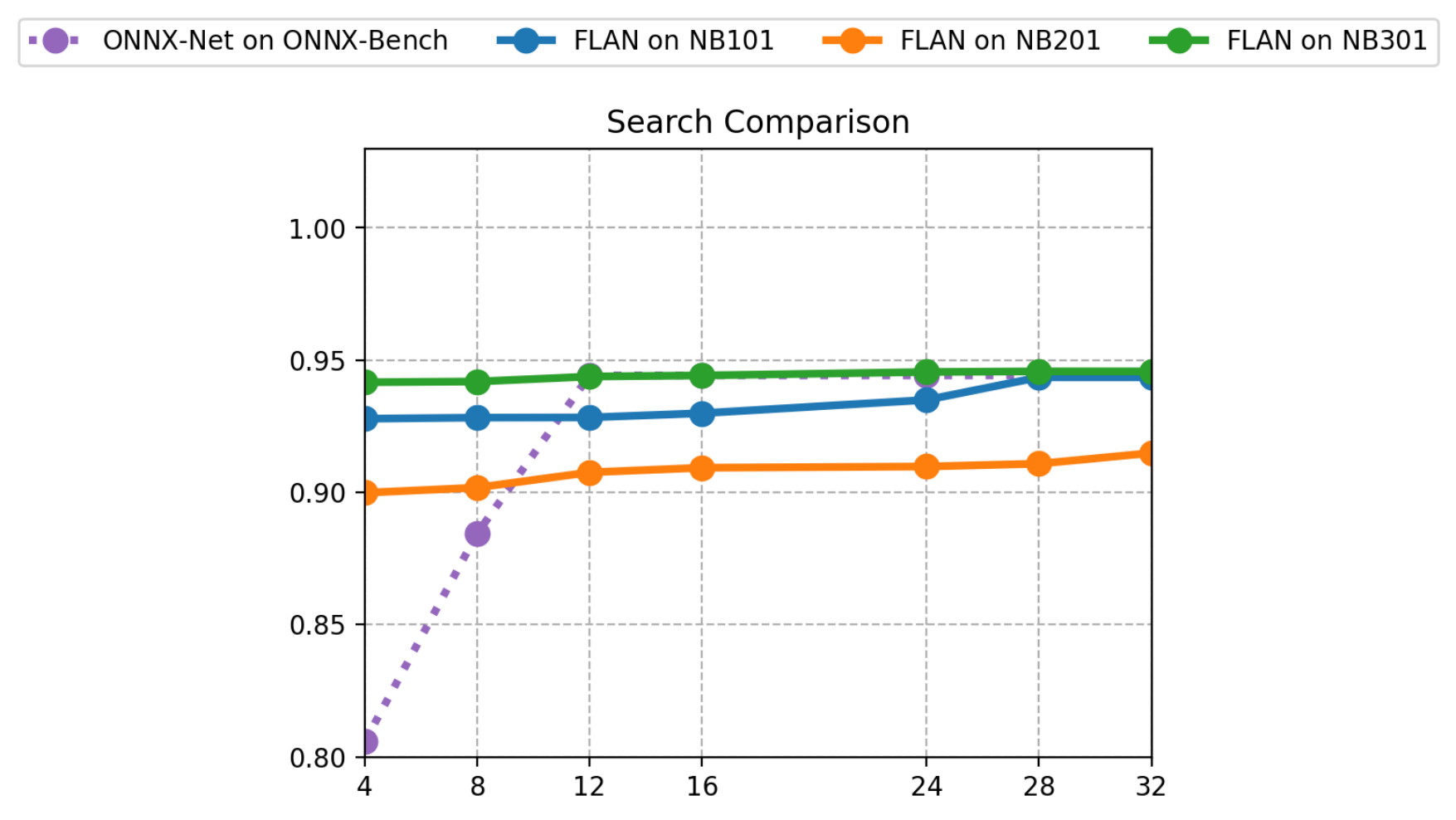}
    \caption{Performance of our ONNX-Net approach using the FLAN-search approach on ONNX-Bench. For comparison, the performance of FLAN on different search spaces is also shown.}
    \label{fig:apdx_search_onnxbench}
\end{figure*}

This section provides additional architecture-search results. Tables~\ref{tab:apdx_search_nb201} and~\ref{tab:apdx_search_onnxbench} report the best test accuracy obtained as the number of trained architectures increases, while Figures~\ref{fig:apdx_search_nb201} and~\ref{fig:apdx_search_onnxbench} visualise the corresponding search trajectories and comparisons with FLAN-based baselines.

\begin{table}[h]
\centering
\captionsetup{skip=2pt}
\caption{Accuracy vs. trained samples for the FLAN-based search of our ONNX-Net approach on the NASBench-201 search space.}
\begin{tabular}{lrrrrrrr}
\toprule
4  &  8  &  12 &  16 &  24 &  28 &  32 \\
\midrule
87.18 & 89.99 & 90.89 & 90.89 & 90.89 & 91.32 & 91.32 \\
\bottomrule
\end{tabular}
\label{tab:apdx_search_nb201}
\end{table}

\begin{table}[h]
\centering
\captionsetup{skip=2pt}
\caption{Accuracy vs. trained samples for the FLAN-based search of our ONNX-Net approach on the ONNX-Bench search space.}
\begin{tabular}{lrrrrrrr}
\toprule
4  &  8  &  12 &  16 &  24 &  28 &  32 \\
\midrule
80.57 &  88.44 &  94.42 &  94.42 &  94.42 &  94.42 &  94.53 \\
\bottomrule
\end{tabular}
\label{tab:apdx_search_onnxbench}
\end{table}

\section{Implementation Details} \label{sec:implementation}

\subsection{Random Seeds}

We use random seed 42 through 46 for all our experiments on multiple seeds. The random seed is used for both training sample selection and training itself.

\subsection{Training Hyperparameters}

The hyperparameters used for training is listed in \cref{tab:hyperparam}.

\begin{table}[h]
\centering
\captionsetup{skip=8pt}
\caption{Hyperparameters used in main table experiments.}
\begin{tabular}{llll}
\toprule
Hyperparameter           & Value               & Hyperparameter    & Value \\ \midrule
Learning Rate            & 5e-5                & Weight Decay      & 0.1   \\
Number of Epochs         & 5                   & Batch Size        & 16    \\
Learning Rate Scheduling & Polynomial          & End Learning Rate & 5e-6  \\
Gradient Accumulation    & 1                   & Warm-up Ratio      & 0.06  \\
Loss Type                & Pairwise Hinge Loss & BF16              & True  \\ \bottomrule
\end{tabular}
\label{tab:hyperparam}
\end{table}

\subsection{Compute Resources}

All our experiments ran on clusters with the following infrastructure:

\begin{itemize}
    \item AMD EPYC 7552 48-Core Processor with 1000GB RAM and 8 × NVIDIA RTX A5500 with 24GB of memory
    \item AMD EPYC 7262 8-Core Processor with 125GB RAM and 7 × NVIDIA A100 with 40GB of memory
\end{itemize}

\begin{table}[t]
\centering
\footnotesize
\captionsetup{skip=2pt}
\caption{Zero-shot transfer learning among einspace, NAS-Bench-101 and NAS-Bench-201. Avg. Kendall's $\tau$ over 5 random seeds is reported.}
\label{tab:zero-shot einspace kendall}
\begin{tabular}{cccc}
\toprule
\multirow{2}{*}{Source → Target Search Space} & \multicolumn{3}{c}{Train Size} \\
           & 200         & 1000        & 5000        \\ \cmidrule(lr){1-1}\cmidrule(lr){2-4}
NAS-Bench-101 → einspace & 0.264 & 0.199 & 0.155 \\
einspace → NAS-Bench-101 & 0.310 & 0.351 & 0.348 \\
einspace → NAS-Bench-201 & 0.242 & 0.198 & 0.180 \\ \bottomrule
\end{tabular}
\end{table}

\begin{table}[t]
\centering
\footnotesize
\captionsetup{skip=2pt}
\caption{Zero-shot transfer learning between NAS-Bench-101 and NAS-Bench-201 using different text encodings. Avg. Kendall's $\tau$ over 5 random seeds is reported.}
\label{tab:encoding ablation kendall}
\begin{tabular}{lccccccc}
\toprule
\multirow{2}{*}{Encoding} & \multicolumn{4}{c}{NB101 → NB201}                         & \multicolumn{3}{c}{NB201 → NB101} \\ 
                          & 200            & 1000           & 5000           & 50000          & 200          & 1000         & 5000        \\ \cmidrule(lr){1-1}\cmidrule(lr){2-5}\cmidrule(lr){6-8} \addlinespace[0.5ex]
Base                      & 0.457          & 0.475          & 0.521          & 0.493          & 0.453        & 0.510        & 0.529       \\ \cmidrule(lr){1-1}\cmidrule(lr){2-5}\cmidrule(lr){6-8} \addlinespace[0.5ex]
+Inputs                   & \textbf{0.566} & \underline{0.571} & \underline{0.574} & \underline{0.560} & 0.509  & \textbf{0.577}        & \underline{0.606}       \\
+Parameters               & 0.549          & 0.542          & 0.536          & 0.536          & \textbf{0.540} & 0.543        & 0.581       \\
+Out Shape                & 0.496          & 0.509          & 0.523          & 0.511          & 0.499        & 0.524        & 0.513       \\ \cmidrule(lr){1-1}\cmidrule(lr){2-5}\cmidrule(lr){6-8} \addlinespace[0.5ex]
Full                      & \textbf{0.566} & \textbf{0.574} & \textbf{0.581} & \textbf{0.571} & \underline{0.539} & \underline{0.564}        & \textbf{0.608}       \\ \bottomrule
\end{tabular}
\end{table}

\begin{table}[t]
\centering
\footnotesize
\captionsetup{skip=2pt}
\caption{Zero-shot transfer learning from NAS-Bench-101 to NAS-Bench-201 using different LM backbone. Avg. Kendall's $\tau$ over 5 random seeds is reported.}
\label{tab:model ablation kendall}
\begin{tabular}{lccccc}
\toprule
\multirow{2}{*}{Model} & \multirow{2}{*}{Model Size} & \multicolumn{4}{c}{NB101 → NB201} \\
                       &                               & 200       & 1000       & 5000      & 50000      \\ \cmidrule(lr){1-1}\cmidrule(lr){2-2}\cmidrule(lr){3-6} \addlinespace[0.5ex]
ModernBERT-base        & 150M                          & \underline{0.548}     & \underline{0.552}      & \underline{0.566}     &   0.559    \\
ModernBERT-large       & 396M                          & \textbf{0.566}     & \textbf{0.574}      & \textbf{0.581}     &   \textbf{0.571}    \\ \cmidrule(lr){1-1}\cmidrule(lr){2-2}\cmidrule(lr){3-6} \addlinespace[0.5ex]
Qwen3             & 752M                         & 0.456     & 0.520      & 0.557     &   \underline{0.566}    \\
Qwen3             & 2.03B                         & 0.489     & \underline{0.552}      & 0.551     &   0.551    \\ \bottomrule
\end{tabular}
\end{table}

\newpage
\section{Graph optimisations} \label{sec:graph optim}
\begin{figure}[h]
\centering
    \begin{tabular}{cc}
        \includegraphics[height=15cm]{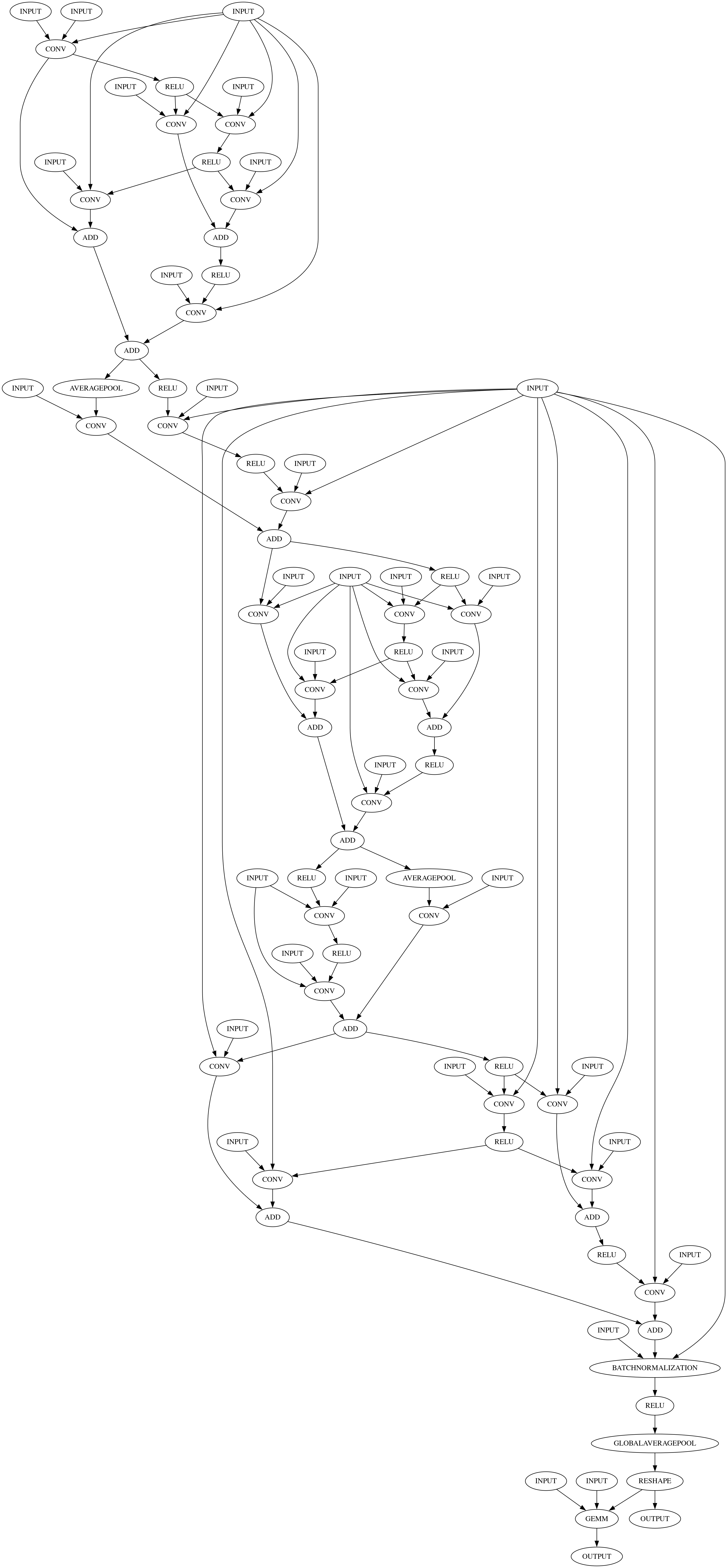}
        &
        \includegraphics[height=15cm]{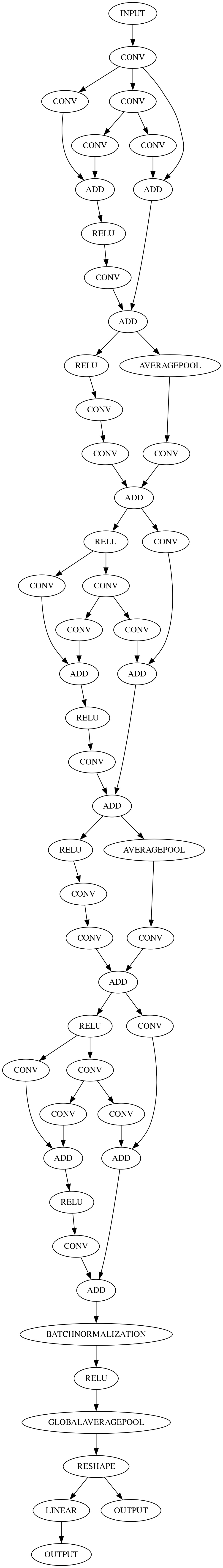}
    \end{tabular}
    \caption{Visualisation of the graph optimisation for a simple neural network. The ONNX graph on the left is simplified in a lossless fashion into the graph on the right.}
    \label{fig:graph_opt_viz}
\end{figure}
\begin{figure}[h]
    \centering
    \includegraphics[width=0.6\linewidth]{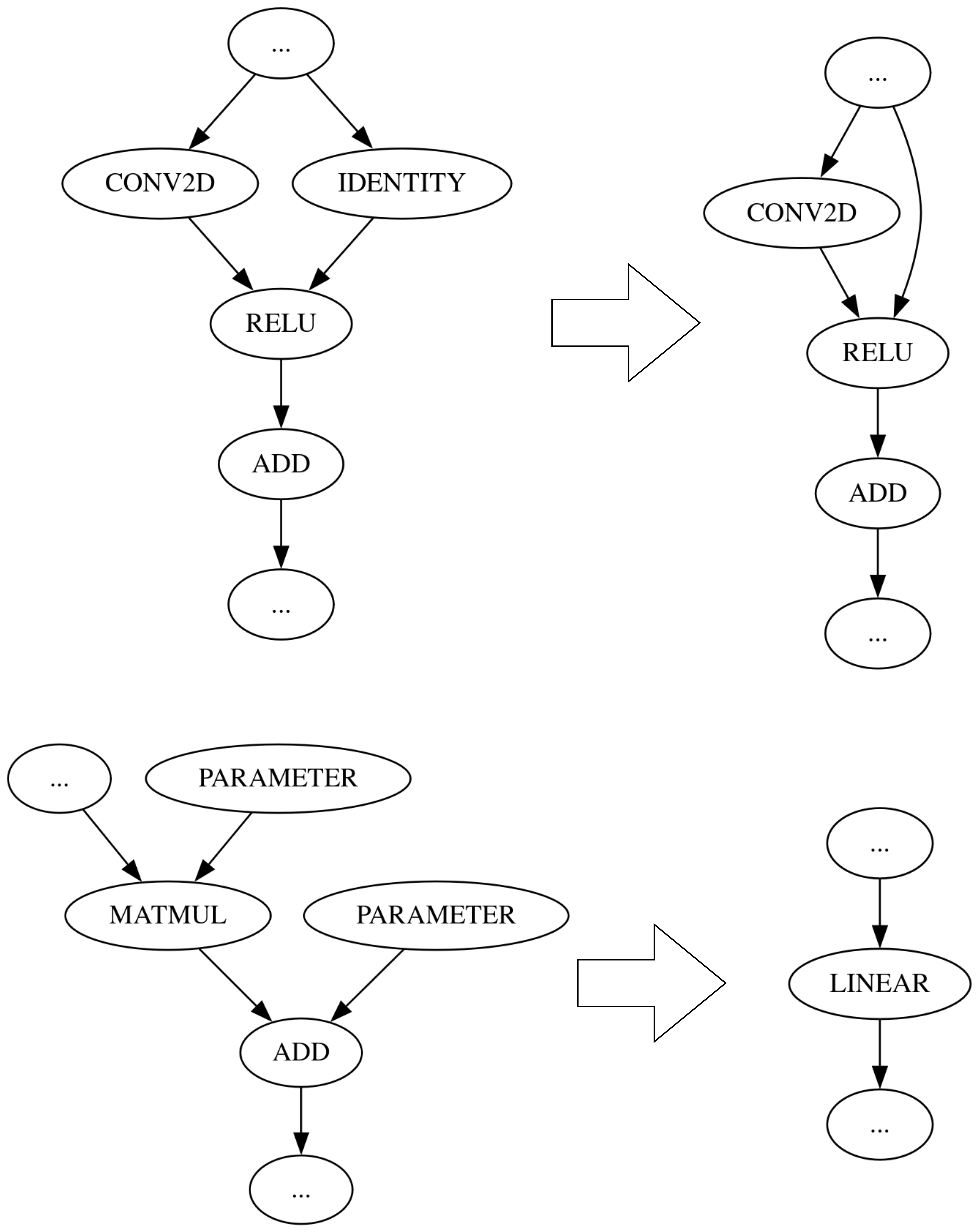}
    \caption{Visualisation of different graph optimisation steps.}
    \label{fig:graph_opt_steps_viz}
\end{figure}

\section{String encoding example} \label{sec:string encoding}
\begin{figure}[h]
\centering
\begin{minipage}{\linewidth}
\begin{verbatim}
Conv(1x3x32x32, 128x3x3x3, 128)(dilations=1,kernel_shape=3,pads=1,strides=1) 
--> Value1:1x128x32x32
Relu(Value1) --> Conv(prev, 32x128x1x1, 32)(dilations=1,kernel_shape=1,pads=0,
strides=1) --> Relu(prev) --> MaxPool(prev)(kernel_shape=3,pads=1,strides=1) 
--> MaxPool(prev)(kernel_shape=3,pads=1,strides=1) --> Value2:1x32x32x32
Concat(Value2, Value2, Value2, Value2) --> Value3:1x128x32x32
Conv(Value3, 32x128x1x1, 32)(dilations=1,kernel_shape=1,pads=0,strides=1) 
--> Relu(prev) --> MaxPool(prev)(kernel_shape=3,pads=1,strides=1) 
--> MaxPool(prev)(kernel_shape=3,pads=1,strides=1) --> Value4:1x32x32x32
Concat(Value4, Value4, Value4, Value4) --> Value5:1x128x32x32
Conv(Value5, 32x128x1x1, 32)(dilations=1,kernel_shape=1,pads=0,strides=1) 
--> Relu(prev) --> MaxPool(prev)(kernel_shape=3,pads=1,strides=1) 
--> MaxPool(prev)(kernel_shape=3,pads=1,strides=1) --> Value6:1x32x32x32
Concat(Value6, Value6, Value6, Value6) --> Value7:1x128x32x32
MaxPool(Value7)(kernel_shape=2,pads=0,strides=2) --> 
Conv(prev, 64x128x1x1, 64)(dilations=1,kernel_shape=1,pads=0,strides=1) 
--> Relu(prev) --> MaxPool(prev)(kernel_shape=3,pads=1,strides=1) 
--> MaxPool(prev)(kernel_shape=3,pads=1,strides=1) --> Value8:1x64x16x16
Concat(Value8, Value8, Value8, Value8) --> Value9:1x256x16x16
Conv(Value9, 64x256x1x1, 64)(dilations=1,kernel_shape=1,pads=0,strides=1) 
--> Relu(prev) --> MaxPool(prev)(kernel_shape=3,pads=1,strides=1)
--> MaxPool(prev)(kernel_shape=3,pads=1,strides=1) --> Value10:1x64x16x16
Concat(Value10, Value10, Value10, Value10) --> Value11:1x256x16x16
Conv(Value11, 64x256x1x1, 64)(dilations=1,kernel_shape=1,pads=0,strides=1) 
--> Relu(prev) --> MaxPool(prev)(kernel_shape=3,pads=1,strides=1) 
--> MaxPool(prev)(kernel_shape=3,pads=1,strides=1) --> Value12:1x64x16x16
Concat(Value12, Value12, Value12, Value12) --> Value13:1x256x16x16
MaxPool(Value13)(kernel_shape=2,pads=0,strides=2) 
--> Conv(prev, 128x256x1x1, 128)(dilations=1,kernel_shape=1,pads=0,strides=1) 
--> Relu(prev) --> MaxPool(prev)(kernel_shape=3,pads=1,strides=1) 
--> MaxPool(prev)(kernel_shape=3,pads=1,strides=1) --> Value14:1x128x8x8
Concat(Value14, Value14, Value14, Value14) --> Value15:1x512x8x8
Conv(Value15, 128x512x1x1, 128)(dilations=1,kernel_shape=1,pads=0,strides=1) 
--> Relu(prev) --> MaxPool(prev)(kernel_shape=3,pads=1,strides=1) 
--> MaxPool(prev)(kernel_shape=3,pads=1,strides=1) --> Value16:1x128x8x8
Concat(Value16, Value16, Value16, Value16) --> Value17:1x512x8x8
Conv(Value17, 128x512x1x1, 128)(dilations=1,kernel_shape=1,pads=0,strides=1) 
--> Relu(prev) --> MaxPool(prev)(kernel_shape=3,pads=1,strides=1) 
--> MaxPool(prev)(kernel_shape=3,pads=1,strides=1) --> Value18:1x128x8x8
Concat(Value18, Value18, Value18, Value18) --> Value19:1x512x8x8
ReduceMean(Value19)(axes=[2,3]) --> Gemm(prev, 10x512, 10) --> Out
\end{verbatim}
\end{minipage}
\caption{A simple neural network in the text representation of ONNX-Net.}
\label{fig:onnx-net-repr}
\end{figure}

\end{document}